\documentclass[a4paper,conference]{IEEEtran}

\ifCLASSINFOpdf
\else
\fi

\usepackage[utf8]{inputenc} % allow utf-8 input
\usepackage[T1]{fontenc}    % use 8-bit T1 fonts
\usepackage{hyperref}       % hyperlinks
\usepackage{url}            % simple URL typesetting
\usepackage{booktabs}       % professional-quality tables
\usepackage{amsfonts}       % blackboard math symbols
\usepackage{nicefrac}       % compact symbols for 1/2, etc.
\usepackage{microtype}      % microtypography

\usepackage{graphicx}
\graphicspath{{./}}
\usepackage{bm}
\usepackage{amsmath}
\usepackage{color}
\usepackage{colortbl}
\usepackage{multirow}
\newcommand{\tabincell}[2]{\begin{tabular}{@{}#1@{}}#2\end{tabular}}
\definecolor{tableColor}{gray}{0.9}
\definecolor{tableRowColor}{gray}{0.9}
\definecolor{ourColor}{gray}{0.97}

\title{Learning Fixation Point Strategy \\for Object Detection and Classification}

\begin{document}
% \title{Bare Demo of IEEEtran.cls\\ for IEEE Conferences}

\author{\IEEEauthorblockN{Jie Lyu}
\IEEEauthorblockA{Xi'an Jiaotong University\\
Email: jiejielyu@outlook.com}
\and
\IEEEauthorblockN{Zejian Yuan}
\IEEEauthorblockA{Xi'an Jiaotong University\\
Email: yuan.ze.jian@xjtu.edu.cn}
\and
\IEEEauthorblockN{Dapeng Chen}
\IEEEauthorblockA{Xi'an Jiaotong University\\
Email: dapengchenxjtu@foxmail.com}
}

\maketitle

\begin{abstract}
We propose a novel recurrent attentional structure to localize and recognize objects jointly. The network can learn to extract a sequence of local observations with detailed appearance and rough context, instead of sliding windows or convolutions on the entire image. Meanwhile, those observations are fused to complete detection and classification tasks. On training, we present a hybrid loss function to learn the parameters of the multi-task network end-to-end. Particularly, the combination of stochastic and object-awareness strategy, named SA, can select more abundant context and ensure the last fixation close to the object. In addition, we build a real-world dataset to verify the capacity of our method in detecting the object of interest including those small ones. Our method can predict a precise bounding box on an image, and achieve high speed on large images without pooling operations. Experimental results indicate that the proposed method can mine effective context by several local observations. Moreover, the precision and speed are easily improved by changing the number of recurrent steps. \emph{Finally, we will open the source code of our proposed approach}.
\end{abstract}

% \fi
\IEEEpeerreviewmaketitle

%% The drawbacks of CNN
%One disadvantage of convolutional networks is that, because they are not recurrent, they rely on hand specified kernel sizes to introduce context. Another disadvantage is that they do not scale well to large images. For example, sequences of handwritten digits must be presegmented into individual characters before they can be recognised by convolutional networks (LeCun et al., 1998a).
\section{Introduction}
%\vspace{-0.3cm}
% the drawbacks of existing methods
Most conventional object detections, such as cascade detector \cite{ViolaJ2004} and deformable part-based models \cite{FelzenszwalbGM2010, SadeghiF2014}, adopt the sliding window strategy, which means classifiers are applied on all possible locations over the entire image.
They are not efficient obviously because the computation resource is distributed equally among the salient regions and the vast majority of background regions.
Recently, neural network methods have achieved great success in object detection. Region proposal based methods \cite{GirshickDD2014,Girshick2015,RenHGS2015} first extract many salient regions, from which the bounding boxes are predicted. But too many proposals slow down the detector. Moreover, the precision is limited by the performance of region proposal methods. 
Single-shot approaches \cite{RedmonDG2016,LiuAE2016} has achieved the state-of-the-art precision and speed, utilizing regression and grid strategies directly. However, it is difficult to detect small objects or those with unusual aspect ratios. Besides, the precision of one-shot process cannot be improved by easily recurrent operations, while the simple strategy of glimpsing more steps is adopted by human to obtain more accurate locations. 

% The mechanism of huamn retinas and the relationship to our method
Human localizes and recognizes an object by several glimpses, each of which contains only details of a small region and rough contextual information\cite{SchmidhuberH1991}, instead of regularly scanning. 
% Context
In addition, contextual information is significant to efficiently localize and recognize the objects in real-world environment \cite{TorralbaQC2006,OlivaT2007}. Especially, the detection of the small object lacking appearance information depends more on the context \cite{BellLB2016}, such as a far and forward car on the highway can be localized by the sky, the ground, the lane lines, and other cars etc. Multi-fixation glimpses are an effective and flexible way to model the relationships between the objects and their context. 

\begin{figure*}[tbp]
	\centering
	\setlength{\belowcaptionskip}{-0.4cm}
	\includegraphics[width=1.0\linewidth]{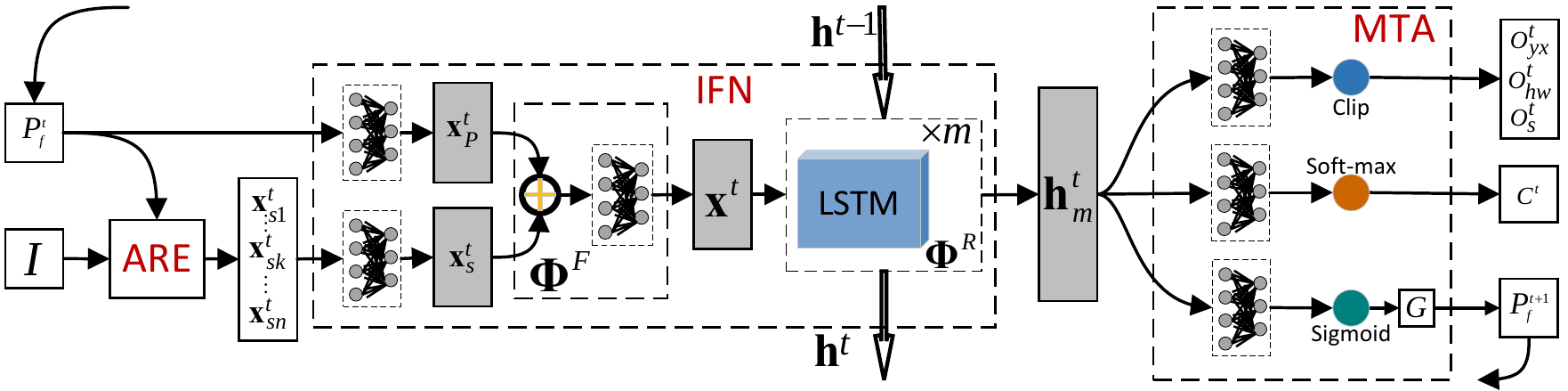}
	\caption{The overview structure of the Recurrent Attention to Detection and Classification Network (RADCN). ARE represents the Attentional Representation Extractor, which utilizes the image $I$ and fixation point $P_f^t$ to extract appearance features. Information Fusion Network (IFN) fuses those multi-scale appearance features and location information, and those local observations over multiple fixations. The output ${\bf{h}}_m^t$ of IFN is fed to multi-task actions (MTA): object detection $[O_{yx}^t, O_{hw}^t, O_s^t]$, classification $C^t$ and the prediction of the next fixation $P_f^{t+1}$.}
	\label{fig_framework}
\end{figure*}

% The introduction of our method
We thus propose a novel structure named Recurrent Attention to Detection and Classification Network shown in Fig.\ref{fig_framework}. It detects and classifies the object by recurrent glimpses, and contains three modules at each step: attentional representation extractor (ARE), information fusion network (IFN) and multi-task actions (MTA). 
Given the input image and the fixation point, ARE mimics the retina to extract multi-scale local regions of detailed appearance and rough context. Following this, it applies Convolutional Neural Network (CNN) learners, acting as the V1 area of the brain, to learn effective appearance representations. 
In IFN, appearance and attentional location information are fused to form a local observation by fully-connected layers. Then 3-layer LSTMs are used to simulate the memory ability of the cerebral cortex, which can fuse multi-fixation local observations and generate a vector of hidden states. 
MTA receives the vector and executes three tasks: the detection and classification of the object, and the prediction of the next fixation point. These tasks are finished by fully-connected layers and limitations of the value range. 

On training, we propose a multi-task loss function to optimize the network end-to-end, and combine stochastic and object-awareness (SA) strategy for learning fixation prediction. 
The stochastic strategy (S) samples the next fixation randomly over a distribution, which enlarges the diversity of samples, and present better results as early as possible by policy-reward method. 
The object-awareness strategy (A) means the last fixation close to the object under L2 criterion, which is beneficial to the stability of optimization and prediction.
Besides, we build a real-world dataset named FCAR, which provides annotations of the forward cars, and can be utilized to train and evaluate the detector of object of interest. 
\emph{We will open the source code of our proposed method}.

\section{Related Work}
%\vspace{-0.2cm}
% Deep Object detection
Deep learning methods are popular in object detection, and have achieved great success recently. 
To avoid sliding windows, Girshick et al. proposed to select salient regions and classified them via CNN \cite{GirshickDD2014}, and the region-based method was improved in terms of speed and precision later \cite{Girshick2015,HeZR2014,RenHGS2015}. Nevertheless, the speed is still slow due to huge amount of regions, and the precision is limited by the performance of region proposal methods. 
One-shot approaches take advantages of regression and grid strategies to directly predict the locations and categories of objects, which is faster than the region-based methods, and achieves higher precision \cite{RedmonDG2016,LiuAE2016}. But they are difficult in locating the small objects without making good use of contextual information. Furthermore, the rough location leads to the instability of object detection and classification.

% Context, Fovea, stochastic latent variable 
The significance of context to localize and recognize objects has been researched in neural computing and cognitive science\cite{TorralbaQC2006,OlivaT2007}. 
Bell et al. explained that context and multi-scale representations provides great improvement in detecting small objects \cite{BellLB2016}. To rationally use computation resource and effectively extract context clues, Larochelle et al. proposed fovea-like image capture \cite{LarochelleH2010}, which is similar to the fovea but too complicate to be used practically. 
Besides, stochastic latent variables are utilized in neural network by Tang et al. and Rezende et al. \cite{TangS2013,RezendeMW2014}, which can imitate the fixation strategy of human attention. Other researchers optimized the network with stochastic latent variables by reinforcement learning methods \cite{Shalev-shwartzBC2016, MnihHG2014}.

% Deep learning and recurrent network, attention
%\cite{Ghahramani2015}\cite{KrizhevskySH2012}
More complex recurrent networks are applied to simulate the attention of the brain \cite{GravesWR2016}. Lots of researches used recurrent or multi-stage networks to learn the attentional position \cite{Ranzato2014,DenilBL2012,XuBK2015,BazzaniLM2011}. 
Mnih et al. proposed Recurrent Attention Model (RAM) and presented the stochastic fixation strategy \cite{MnihHG2014}. Deep Recurrent Attentive Writer (DRAW) is proposed by Gregor et al. to learn the latent distribution of samples, and they also apply DRAW to finish classification task \cite{GregorDG2015}. While none of them detects and classifies the objects jointly by recurrent network, and the hybrid strategy of stochastic and object-awareness fixation has not been discussed. 
Caicedo et al. utilized reinforcement learning method to localize objects with 9-action policies \cite{CaicedoL2015}, but frequently resizing cause too much deformation of objects and the location is not precise, meanwhile, repeatedly processing the large image makes the speed slow. 

\section{Recurrent Attention to Detection and Classification Network}
\vspace{-0.3cm}
We propose a novel recurrent network to jointly predict the category and bounding box of an object with related local observations. 
The overview structure named RADCN is illustrated in Fig. \ref{fig_framework}. 
Three main modules in the network are: Attentional Representations Extractor (ARE), Information Fusion Network (IFN) and Multi-task Actions (MTA).
\begin{figure}[b]
	\centering
	\setlength{\belowcaptionskip}{-0.2cm}
	\includegraphics[width=0.9\linewidth]{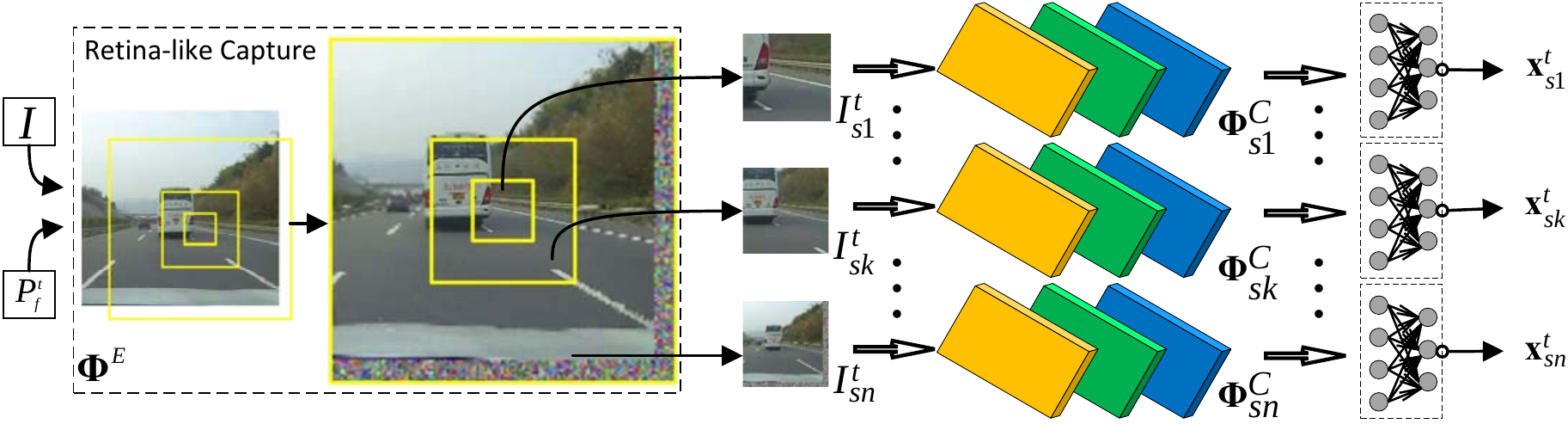}
	\caption{The network of Attentional Representation Extractor (ARE). The ARE consists of two components: (1) The retina-like capture, which obtains multi-scale patches from an image $I$ and attentional point $P_{f}^{t}$. (2) Several CNN learners, which are applied to learn the representations on each scaled patch $I_{sk}^{t}$.}
	\label{fig_are}
\end{figure}
\subsection{Attentional Representations Extractor}
%\vspace{-0.2cm}
A rich collection of contextual associations is important to search objects in the real-world environment \cite{OlivaT2007}.
Simulating the retina and the V1 area, we joint the retina-like capture and CNN learners together, which can provide attentional representations with abundant contextual information and significant spatial structures. 
The network of the proposed Attentional Representation Extractor is shown in Fig.\ref{fig_are}.

{\bf{Retina-like Capture}}: Retina-like capture \cite{SchmidhuberH1991} ${\bm{\Phi}}^{E}$ is the first module of our proposed network, which controls the source of appearance information flow. Centering on the given fixation point $P_{f}^{t}$, it generates several scaled patches $I_{sk}^{t}, k=1,2,\dots, n,$ from an entire image $I$ at $t$ step with specific parameters ${\bm{\theta}}_{E}$. Given the number of scales $n$, the size of the minimum scale $s_0$ and the scaled factor $s_f$, we crop $k$-th patches with size $s_k = s_0\times s_f^{k-1}$, and these patches are resized into an unified size $s_u$. Mathematically, the mapping ${\bm{\Phi}}^{E}$ can be presented as
\begin{equation}
	\setlength{\abovedisplayskip}{1ex} 
	\setlength{\belowdisplayskip}{1ex}
	\{I_{sk}^{t}\} = {\bm{\Phi}}^{E}(I, P_{f}^{t}, {\bm{\theta}}_{E}), k=1,2,\dots,n,
	\label{equ_retina_like}
\end{equation} 
in which ${\bm{\theta}}_{E}=\{s_0, n, s_f, s_u\}$ is hand-crafted by a little priory knowledge of the input size. The fixation point at the first step is sampled from an uniform distribution.

{\bf{CNN learners}}: A group of CNN learners are applied to learn effective representations on those scaled patches $\{I_{sk}^{t}\}, k=1,2,\dots,n$. Each CNN learner consists of three convolutional layers ${\bm{\Phi}}_{sk}^{C}$ with parameter set ${\bm{\theta}}_{sk}^{C}$ and a fully-connected layer with weight ${\bf{W}}_{sk}^{C}$ and bias ${\bf{b}}_{sk}^{C}$, where the convolutional layer involves convolution operation, max-pooling(except for the second layer) and ReLU activation. The output of the $k$-th CNN learner is
\begin{equation}
	\setlength{\abovedisplayskip}{1ex} 
	\setlength{\belowdisplayskip}{1ex}
	{\bf{x}}_{sk}^{t} = \max\{{\bf{W}}_{sk}^{C}{\bm{\Phi}}_{sk}^{C}(I_{sk}^{t}, {\bm{\theta}}_{sk}^{C})+{\bf{b}}_{sk}^{C}, {\bf{0}}\}.
\end{equation}
The parameters of CNN learners can be trained for hybrid task of detection and classification, which is a flexible way to fit various errands.

%\vspace{-0.2cm}
\subsection{Information Fusion Network}
%\vspace{-0.2cm}
Similar to human visual system, the position of fixation is fused with appearance features in our proposed method. We concatenate attentional representations $\{{\bf{x}}_{sk}^{t}\}$ of multiple scales and map them into an appearance descriptor ${\bf{x}}_{s}^{t}$ via fully-connected layer. Meanwhile, the coordinates of fixation point $P_{f}^{t}$ are also projected into a location descriptor ${\bf{x}}_{P}^{t}$ with the same dimension as ${\bf{x}}_{s}^{t}$. 

{\bf{Appearance \& Location Fusion}}: Several operations following by fully-connected layer can be utilized to fuse appearance and location descriptors, such as addition ${\bm{\Phi}}_{+}^{F}$, element-wise multiplication ${\bm{\Phi}}_{\times}^{F}$ and concatenation ${\bm{\Phi}}_{c}^{F}$ etc. Furthermore, the fusion descriptor ${\bf{x}}^{t}$ is formulated as 
\begin{equation}
	\setlength{\abovedisplayskip}{1ex} 
	\setlength{\belowdisplayskip}{1ex}
	{\bf{x}}^{t} = {\bm{\Phi}}^{F}({\bf{x}}_{s}^{t}, {\bf{x}}_{P}^{t}, {\bm{\theta}}^{F}), {\bm{\Phi}}^{F}\in\{{\bm{\Phi}}_{+}^{F}, {\bm{\Phi}}_{\times}^{F}, {\bm{\Phi}}_{c}^{F}\},
\end{equation} 
in which ${\bm{\theta}}^{F}$ is the set of parameters including the weight and bias of the fully-connected layer.

{\bf{Multi-Fixation Fusion}}: It is essential to joint these local observation vectors $\{{\bf{x}}^{t}\}, t=1,2,\dots,T,$ over multiple steps, where $T$ is the number of fixations. 
We thus take $m$-layer LSTMs to learn the dependencies and contributions of these observations $\{{\bf{x}}^{t}\}$, which performs similarly to the cerebral cortex remembering lots of general rules. 
The hidden states of the last layer ${\bf{h}_{m}^{t}}$ is taken as the output of LSTMs ${\bm{\Phi}}^{R}$. Specifically, the fusion of multiple steps is represented as
\begin{equation}
	\setlength{\abovedisplayskip}{1ex} 
	\setlength{\belowdisplayskip}{1ex}
	{\bf{h}}_{m}^{t} = {\bm{\Phi}}^{R}({\bf{h}}^{t-1}, {\bf{x}}^{t}, {\bm{\theta}}^{R}), t=1,2,\dots,T,
\end{equation} 
where ${\bm{\theta}}^{R}$ is the parameters, and ${\bf{h}}^{t-1}=\{{\bf{h}}_{1}^{t-1}, {\bf{h}}_{2}^{t-1},\dots,{\bf{h}}_{m}^{t-1}\}$ is the hidden states of all LSTMs at the previous step.
Different from conventional method \cite{SadeghiF2014} and r-CNN detector \cite{Girshick2015}, the multi-fixation fusion provides a flexible way to recognize objects from local to global.

\subsection{Multi-task Actions}
%\vspace{-0.2cm}
Three main tasks of our proposed network are detection, classification, and prediction of the next fixation point, all of which take the multi-fixation fusion vector ${\bf{h}}_{m}^{t}$ as their inputs .

{\bf{Detection}\&\bf{Classification}}: The object geometric parameters $O^t=[O_{yx}^{t}, O_{hw}^{t}, O_{s}^{t}]$ include the left-up position and the size along with a score. They are regressed by a fully-connected layer with parameters ${\bf{W}}_A^D$, ${\bf{b}}_A^D$, and clip operations $S_c$ with super parameters ${\bm{\theta}}_c$ are used to ensure the numerical range. The probabilities of all categories $C^{t}$ is obtained by a fully-connected layer with parameters ${\bf{W}}_A^C$, ${\bf{b}}_A^C$, and normalized by a soft-max operation $S_f$. Specifically, these can be expressed as
\begin{equation}
	\setlength{\abovedisplayskip}{1ex} 
	\setlength{\belowdisplayskip}{1ex}
	\begin{aligned}
		& O^t=S_c({\bf{W}}_{A}^{D}{\bf{h}}_{m}^{t} + {\bf{b}}_A^D, {\bm{\theta}}_c)\\
		& C^t=S_f({\bf{W}}_{A}^{C}{\bf{h}}_{m}^{t} + {\bf{b}}_A^C)
	\end{aligned}
\end{equation}

{\bf{Prediction of Fixation Point}}: We sample the next fixation point from a Gaussian distribution with a mean ${\bm{\mu}}_{P}^{t+1}$ and standard deviate $\sigma$. A fully-connected layer with parameters ${\bf{W}}_A^{P}, {\bf{b}}_A^{P}$ and sigmoid activation are utilized to estimate the mean ${\bm{\mu}}_{P}^{t}$. The sampling process can be presented as
\begin{equation}
	\setlength{\abovedisplayskip}{1ex} 
	\setlength{\belowdisplayskip}{1ex}
	P_{f}^{t+1} \sim N({\bm{\mu}}_P^{t+1}, \sigma^2), {\bm{\mu}}_{P}^{t+1}=2S({\bf{W}}_A^{P}{\bf{h}}_{m}^{t}+{\bf{b}}_A^{P})-1,
\end{equation}
where $N(\cdot, \cdot)$ represents Gaussian distribution, and $S(\cdot)$ is the sigmoid function. Those weighted parameters of the whole network can be trained by a hybrid loss function end-to-end.

\section{Hybrid Loss Function}
To complete multiple tasks, we minimize a hybrid loss function $L_{h}$,
\begin{equation}
	\setlength{\abovedisplayskip}{1ex} 
	\setlength{\belowdisplayskip}{1ex}
	L_{h} = L_{d} + L_{c} + L_{P},
\end{equation}
where $L_{d}$ is the loss of detection, $L_{c}$ denotes the cross-entropy of classification, and $L_{P}$ represents the loss of fixation points.

\subsection{Detection and Classification Loss}
{\bf{Detection}}: 
We suppose that an image $I$ contains only one object of interest with the true left-up position $O_{yx}^{g}$ and size $O_{hw}^{g}$. 
The prediction of the position $O_{yx}^{T}$ at the last step is regarded as a random vector conditioned on the groundtruth, which can be optimized by maximum logarithmic likelihood. In addition, a loss item is required to optimize the prediction of the size $O_{hw}^T$. We take the Intersection of Union (IoU) between the predicted bounding box and the groundtruth to construct the loss related to the size prediction. Experimentally, we limit the range of $IoU(\cdot)$ mapping to be within $[10^{-8}, 1]$.
The loss function $L_d$ of detection task is given
\begin{equation}
	\setlength{\abovedisplayskip}{1ex} 
	\setlength{\belowdisplayskip}{1ex}
	L_d = -\ln(P(O_{yx}^T)) -\ln(IoU(O_{hw}^T, O_{hw}^g, O_{yx}^T, O_{yx}^g)).
\end{equation}
The error of predicted position vector $(O_{yx}^{T} - O_{yx}^{g})$ is selected as a Gaussian function with the zeros mean, namely 
\begin{equation}
	\setlength{\abovedisplayskip}{1ex} 
	\setlength{\belowdisplayskip}{1ex}
	O_{yx}^{T} \sim N(O_{yx}^{g}, \sigma_O^2),
\end{equation}
where $N(\cdot, \cdot)$ represents Gaussian distribution, the $x$ and $y$ coordinates are supposed to be independent, and $\sigma_{O}$ is the standard deviate. 

The maximum likelihood item will guide the adjustment of weighted parameters, if the overlap between the prediction and the groundtruth is zero. While it is non-zero, maximizing the IoU item can improve the precision of the predicted bounding box.

{\bf{Classification}}: 
We use cross-entropy to measure the predicted distribution $C^T$ and the true distribution $C^g$ over categories. The loss of classification task is formulated as $L_c = \sum_{k=1}^{N}-C_k^g\ln(C_k^T),$
where $N$ is the number of categories, $C_k^g$ and $C_k^T$ represent the $k$-th component of $C^g$ and $C^T$ respectively.

\subsection{Fixation Process Loss}
Besides the loss of detection and classification, a loss related to fixation points is also required to constrain the fixation prediction, and efficiently extract more effective local observations. 

Policy-reward mechanism is utilized to optimize a decision process in reinforcement learning. In our method, the policy of fixation is over Gaussian distribution with an optimal mean value. We define the cumulative reward $R^t$ as the overlap rate between the predicted bounding box and the true one by IoU mapping. The decision process of fixation points thus can be optimized by a policy-reward method \cite{Williams1992}, the loss item of which can be represented as
\begin{equation}
	\setlength{\abovedisplayskip}{1ex}
	\setlength{\belowdisplayskip}{1ex}
	L_P^r = -\sum_{t=1}^{T}\ln(\pi(P_{f}^{t}|{\bm{\mu}}_P^{t}, \sigma^2))(R^t-b^t),
\end{equation}
where $\pi(\cdot, \cdot, \cdot)$ is the distribution of sampling point $P_{f}^{t}$, $b^t$ is the expectation of $R^t$\cite{SuttonMS1999}, namely $b^t = E_{\pi}[R^t]$. Minimizing the loss $L_P^r$ can obtain lower variance of gradient estimation \cite{MnihHG2014}. 

Human will focus his attention to the object they intend to recognize. To mimic this mechanism, we takes L2 loss to constrain the mean of the last fixation close to the center of an object. The whole loss of the fixation process is presented as
\begin{equation}
	\setlength{\abovedisplayskip}{1ex}
	\setlength{\belowdisplayskip}{1ex}
	L_P = L_P^r + \frac{1}{T}\sum_{t=1}^{T}(O_s^t-R^t)^2 +  \|{\bm{\mu}}_P^{T}-(O_{yx}^g+\frac{1}{2}O_{hw}^g)\|_2^2,
\end{equation}
where the score $O_s^t$ is optimized to estimate the reward $R^t$ under L2 criteria. The total loss of fixation contains both stochastic and object-awareness strategies, called SA. The stochastic strategy (S) enlarges the diversity of samples and can enhance the generalization ability, while the item $L_P^r$ is used to improve the attentional efficiency. The object-awareness strategy (A) makes the last fixation close to an object, which leads to the stability of the prediction and optimization.

\section{Experiments}
\subsection{Dataset}
%\vspace{-0.2cm}
Our experiments are carried on three datasets, including two variants of the MNIST dataset and a real-world dataset of car detection.

{\bf{MSO Dataset} \& \bf{MSNO Dataset}}: 
We build a new dataset named MNIST Scaled Object (MSO) dataset, on which an image contains a bounding box and a label of the only one digital object. Images of the MSO dataset are generated by two steps:(1) Resize the images of the MNIST dataset by random scales sampled from a uniform distribution over [0.3, 1.5]. (2)Insert the images into a $56\times56$ dark background at random locations, ensuring that the digital numbers are not truncated. To test the sensitivity to noises, we build another dataset set named Mnist Scaled and Noised Object(MSNO) by adding 6 patches of size $6\times6$ into the images of MSO. The noising patches are cropped from corresponding subset of the MNIST dataset randomly. 

{\bf{FCAR Dataset}}:
Particularly, we build a real-world dataset named FCAR dataset for car detection, of which the images may contain more than one car, but the object we are interested in is just the one on the same lane with our car. Therefore, the annotation of an image is a bounding box of the attentional car. The real-world images taken by fixed camera contains more abundant structure and context information of the scene, which is significant for the detection of attentional object, especially those small objects. For the FCAR dataset, the trainset includes $6330$ images with size $800\times800$, and the testset has $2712$ images.

\subsection{Implemented Details}
{\bf{Initial Fixation Point}}: The initial fixation point is essential for the proposed network. We normalize the image coordinates to [-1, 1], and put the original point (0. 0) to the center of the image. We randomly sample the initial point from an uniform distribution over a range $[-r, r], r\in[0, 1]$. 

{\bf{Parameter Settings}}: The setting of the structure parameters of our proposed RADCN are illustrated in Table \ref{tab_param_setting}. The parameters of ARE are adjusted according to the size of the input image, and all the CNN learners have same structure but with different parameters.
\begin{table*}[htbp]
	\setlength{\abovecaptionskip}{-1pt}
	\centering
	\caption{The setting of structure parameters in these subnet of the proposed RADCN on the MSO and MSNO dataset. On the FCAR dataset, we just modify the setting of ARE to $s_0=32\!\times\!32,s_f=4.5,n=3,s_u=32\!\times\!32$, change the first convolution to $5\!\times\!5\!\times\!3\!\times\!32$, remove the classification action, and keep other parameters unchanged.}
	\begin{tabular}{c|c|c|c|c}
		\hline
		SubNet 	& ARE & CNN learners & Fusion Network & Actions\\
		\hline
		Params 	& \tabincell{l}{\small{$s_0=8\!\times\!8$}\\\small{$s_f=2.0$}\\\small{$n=3$}\\\small{$s_u=2s_0$}} & \tabincell{l}{\small{conv1+relu1: $5\!\times\!5\!\times\!1\!\times\!32$}\\\small{conv2+relu2: $5\!\times\!5\!\times\!32\!\times\!64$}\\\small{conv3+relu3: $5\!\times\!5\!\times\!64\!\times\!4$}\\\small{fc4+relu4:$1024\!\times\!256$}\\\small{softmax}} & \tabincell{l}{\small{scale fusion:} \\\small{fc+relu:$768\!\times\!256$}\\\small{location mapping:} \\\small{fc+relu:$2\times256$}\\\small{fusion method: $\bm{\Phi}_+^F$}\\\small{multi-fixation fusion:}\\\small{LSTMs:$[256]\times3$}} & \tabincell{l}{\small{fc+clip:$256\!\times\!2\to O_{yx}^t$}\\\small{fc+clip::$256\!\times\!2\to O_{hw}^t$}\\\small{fc+clip:$256\!\times\!1\to O_s^t$}\\\small{fc+softmax:}\\\small{ $256\!\times\!10\to C^t$}\\\small{fc+sigmoid+$N(\bm{\mu}, \sigma)$:}\\\small{ $256\!\times\!2\to P_{f}^t, \sigma=0.2$ }}\\
		\hline
	\end{tabular}
	\label{tab_param_setting}
\end{table*}
\begin{figure}[htbp]
	\centering
	\setlength{\belowcaptionskip}{-0.2cm}
	\includegraphics[width=1.0\linewidth]{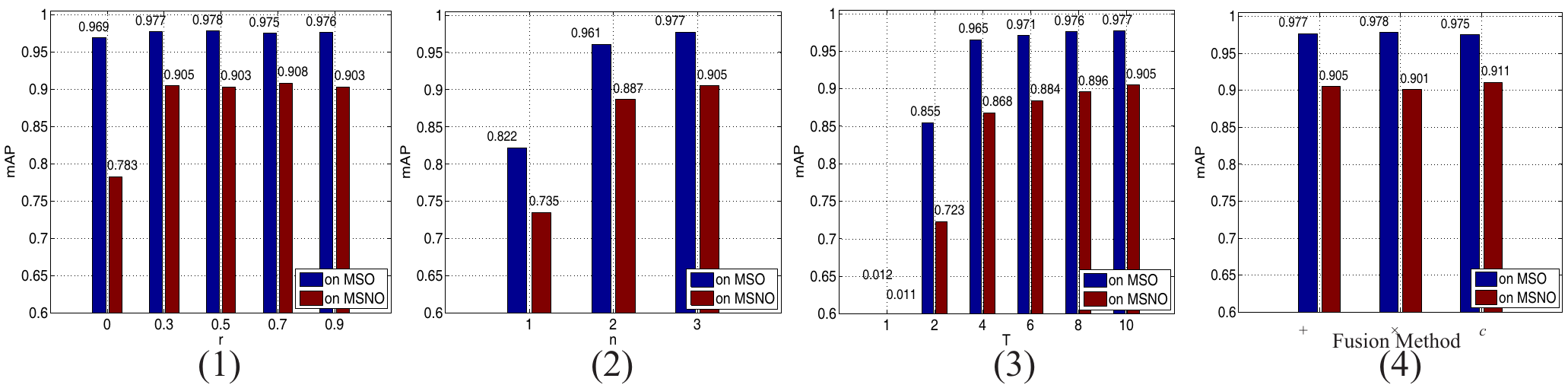}
	\caption{The comparisons of different parameter settings. (1) The mean AP under different initial range; (2) The results when different number of image scales are adopted in ARE; (3) The necessity of multi-fixation glimpses; (4) The comparison of three kinds of fusion methods.}
	\label{fig_param}
\end{figure}

\subsection{Effects of the Structure Parameters}
On MSO and MSNO dataset, we explore the effects of the structure of the proposed RADCN by four groups of experiments, in which the default structure parameters are $r=0.3, n=3, T=10, \bm{\Phi}^F=\bm{\Phi}_+^F$, and each group of experiments change one of these parameters. 

{\bf{Initial range}}: Randomly selecting initial point is a way to augment and enlarge the dataset because of the different local observations. We sample the initial point from an uniform distribution over the range $[-r,r], r\in\{0, 0.3, 0.5, 0.7, 0.9\}$. Fig.\ref{fig_param} (1) shows the mAP values under different ranges. Without randomness ($r=0$), the precision on the MSNO dataset is about $19\%$ less than that on the MSO dataset, which indicates that random initialization is beneficial to learn a model robust to clutter background. \emph{Results with different initial points and fixation paths are shown in Supplementary 2}.

{\bf{Number of scales}}: Multiple scales provide more contextual information which can guide the prediction of the next fixation point. We change the number of scales $n$ from $1$ to $3$ to explore the effects of multi-scale method. Fig.\ref{fig_param} (2) manifests that 3-scale local regions are more effective to contain context, and achieve 0.977 mAP value on the MSO dataset.  

{\bf{Glimpse steps}}: Glimpsing by multiple steps is another way to mine abundant contextual clues, and the number of steps T is chosen from $\{1, 2, 4, 6, 8, 10\}$ in our experiments. As shown in Fig.\ref{fig_param} (3), a random glimpse ($T$=1) is a bad choice. With the growth of glimpses, results are getting better. About $0.97$ mAP value is achieved by the proposed RADCN when $T\ge6$ on the MSO dataset. As a conclusion, multiple steps is an essential way to improve performance. \emph{The attentional processes are illustrated in Supplementary 1}. 

{\bf{Fusion methods}}: Three kinds of methods are provided to fuse appearance and location information, namely addition($+$), production($\times$) and concatenation($c$). Fig.\ref{fig_param} (4) illustrates that the three methods obtain similar mAP values, about $0.97$ on the MSO dataset and $0.90$ on the MSNO dataset. For computational simplicity, we choose the addition as the fusion method.

\subsection{Significance of the Fixation Point Strategy}
Experiments on the MSO and MSNO dataset are carried out to analyze the significance of stochastic and object-awareness strategies. The stochastic strategy (S) is trained by a variance of REINFORCE rule, and the object-awareness one (A) without randomly sampling is constrained by L2 loss with the true location of the object. The hybrid method (SA) of the stochastic and object-awareness method are also taken into consideration in experiments.   
\begin{figure}[htbp]
	\centering
	\setlength{\belowcaptionskip}{-0.2cm}
	\includegraphics[width=1.0\linewidth]{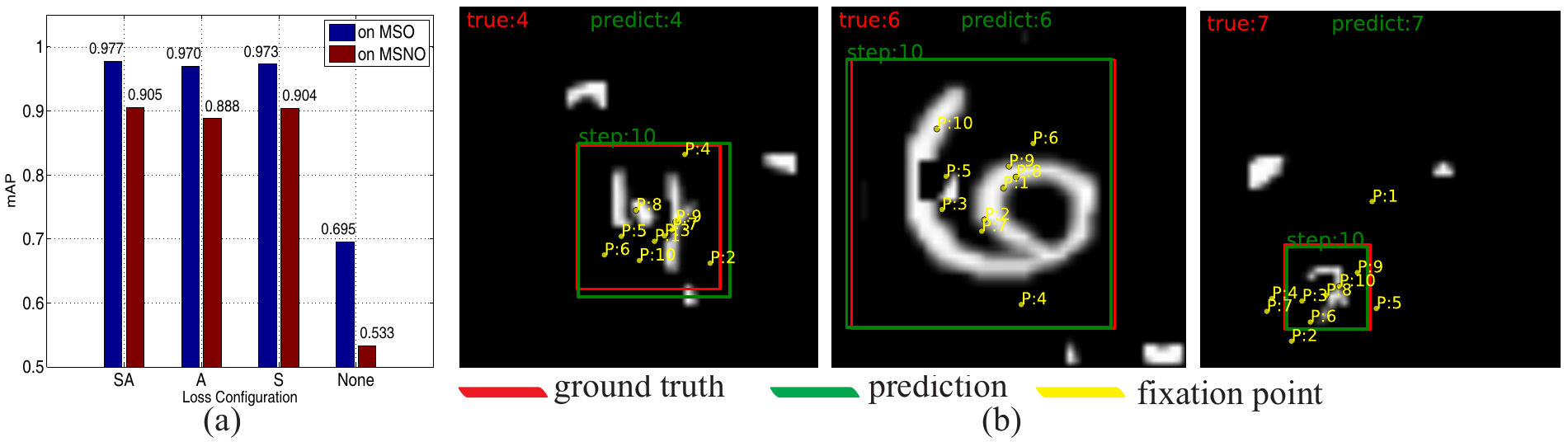}
	\caption{(a) The comparison of different fixation point strategies on MSO dataset and MSNO dataset. S denotes the stochastic strategy and A represents object awareness of the last fixation step. (b) The illustration of several instances obtained by the proposed network on the MSNO dataset. \emph{More results are provided in Supplementary 3}.}
	\label{fig_loss}
\end{figure}

Fig. \ref{fig_loss} (a) indicates that the L2 constraint is effective to guide the object-awareness strategy to predict the next fixation, while the stochastic one is more robust to clutter background, and achieve a gain of more than $1.6$ percentage on the MSNO dataset. However, we also discover that the stochastic method may fail to achieve convergent state occasionally. The hybrid method (SA) can obtain the highest mAP of 0.905, and is very stable in convergence. The strategy (None) without both S and A obtains bad results on the MSO and MSNO dataset, which illustrates the necessity of the stochastic method and the L2 loss. Several instances generated by the hybrid method are illustrated in Fig. \ref{fig_loss} (b). We can see that even the small objects and the corrupted ones can be detected and classified accurately. 

We also compare our RADCN method with other networks, shown as Table \ref{tab_comp_msno}. On the cluttered translated $100\times100$ (CT100) dataset \cite{MnihHG2014}, the results suggest that it is beneficial to jointly localize and recognize objects along with our fixation strategy. On the MSNO dataset, our RADCN achieves the best precision, and the proposed method is better in detecting and classifying the small object with cluttered background. 
\begin{table*}[htbp]
	\centering
	\caption{The comparisons of our method with other existing networks on the MSNO dataset and the cluttered translated $100\times100$(CT100) dataset.}
	\begin{tabular}{c|cc|cc|cc|cc}
		\hline
		Methods 							& \multicolumn{2}{c|}{RADCN(our)} 			& \multicolumn{2}{c|}{LeNet+Regression \cite{LecunBD1989}} 	& \multicolumn{2}{c|}{RAM \cite{MnihHG2014}}		& \multicolumn{2}{c}{DRAW \cite{GregorDG2015}}	\\
		\hline
		Dataset								& MSNO	& CT100								& \multicolumn{2}{c|}{MSNO}			& MSNO	& CT100				& MSNO	& CT100				\\
		\hline
		mean IoU    						& 0.879	&0.918								& \multicolumn{2}{c|}{0.643}		& 	--	& --				&	--	&	--				\\
		\rowcolor{tableRowColor}error rate  & 6.8\% &3.72\%	 							& \multicolumn{2}{c|}{42.1\%}		&19.7\% & 8.11\%			&19.5\%&	3.36\%			\\
		mAP 								& 0.905 &0.940	 							& \multicolumn{2}{c|}{0.395}		& 	--	& --				&	--	&	--				\\
		\hline
	\end{tabular}
	\label{tab_comp_msno}
\end{table*}

\subsection{Attentional Object Detection on Real-world Images}
\vspace{-0.2cm}
We apply the proposed RADCN on the FCAR dataset to detect the forward car which is in the same lane with our car, and evaluate the precision of the detection by mean IoU. For training process, the cross-entropy of classification is removed due to the absence of true label information. 
\begin{figure}[htbp]
	\centering
	\setlength{\belowcaptionskip}{-0.2cm}
	\includegraphics[width=1.0\linewidth]{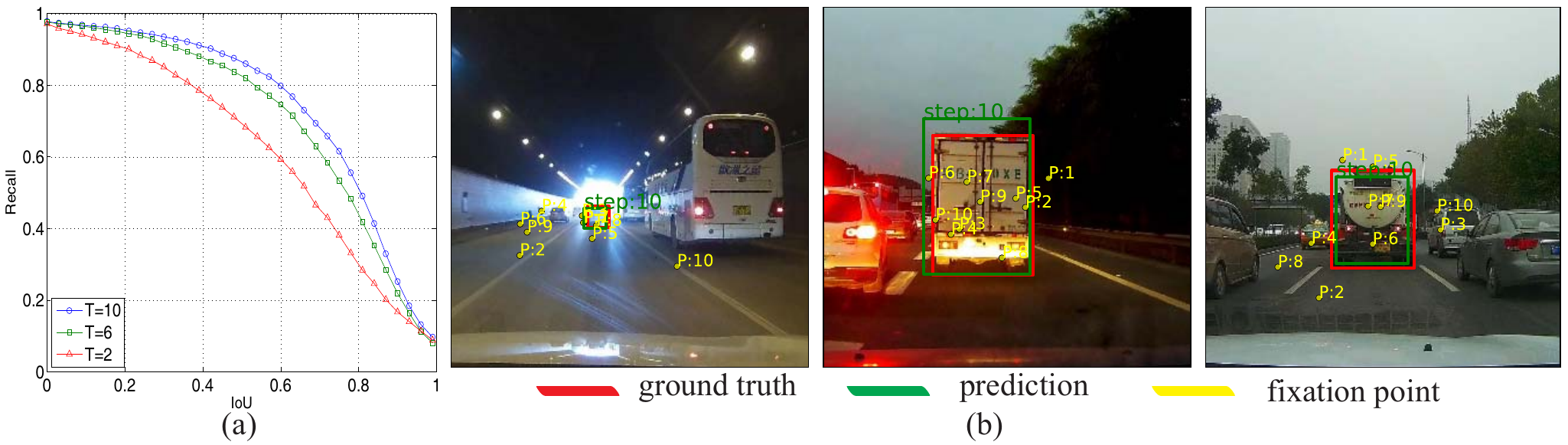}
	\caption{(a) The comparisons of different glimpse steps. (b) The illustration of typical results obtained by the proposed RADCN on the real-world FCAR dataset. \emph{Attentional process and more results are shown in Supplementary 4 and 5}.}
	\label{fig_fcar}
\end{figure}

As shown in Fig. \ref{fig_fcar}(a), increasing the number of glimpse steps is a simple and effective way to improve the precision. Fig. \ref{fig_fcar}(b) illustrates that the proposed RADCN is effective to detect small and large objects, especially, it can ignore other cars in different lanes, and localize the attentional car. The predicted bounding boxes achieve $0.70$ mean IoU. With fixed viewpoint, the images imply the distributions of these objects. Besides, lots of contextual information guides the network to localize the forward car, such as the sky, the ground, lane lines and other cars etc. Therefore, our proposed model is flexible and efficient to learn effective features, mine abundant environmental information, and fuses multi-fixation partial observations to detect the attentional object. 

Compared with the method of convolution on the entire image, the approach of multi-scale local regions is much faster. The local regions are resized to $m\!\times\!m$, while the size of the entire image is $M\!\times\!M, M>>m$. The number of convolutions on the global image is $N_g = \frac{M^2}{nTm^2}N_l$, where $n$ is the number of scales, $T$ is the number of steps, and $N_l$ is the number of convolutions on those local regions. On the FCAR dataset, $M=800, m=32, n=3, T=10$. As a result, $\frac{N_g}{N_l}>20$, which illustrates that method of those local regions costs far less computation than that of the entire image. The proposed RADCN can achieve about 30 fps on $800\times800$ images when $T=10$ on TensorFlow framework and M40 GPU. \emph{We will open the source code of our proposed method}.

% \vspace{-0.2cm}
\section{Discussion}
\vspace{-0.2cm}
Our RADCN is effective to jointly localize and recognize objects efficiently. Furthermore, the hybrid loss function of multiple tasks can be used to learn the parameters of the network end-to-end. Especially, the stochastic fixation and object-awareness strategy makes our network stable and accurate. Experiments on the FCAR dataset demonstrate that our method can extract useful contextual information to detect attentional objects, especially those small objects.\\
Our current model detects only one objects in each image. In the future work, we will extend it to multi-object detection and classification.

\vspace{-0.2cm}	
\bibliographystyle{IEEEtran}
\bibliography{IEEEabrv,egbib}

\end{document}